\documentclass[9pt]{elife}

\usepackage{lipsum} 
\usepackage[version=4]{mhchem}
\usepackage{siunitx}
\DeclareSIUnit\Molar{M}

\usepackage{bm}
\usepackage{graphicx}
\usepackage{svg}
\usepackage{amsthm}
\usepackage{xcolor}
\usepackage{comment}
\usepackage{multirow}
\usepackage{booktabs}
\usepackage{algorithm2e}

\title{A differentiable model for optimizing the genetic drivers of synaptogenesis}

\author[1*]{Tommaso Boccato}
\author[1]{Matteo Ferrante}
\author[1,2]{Nicola Toschi}
\affil[1]{Department of Biomedicine and Prevention, University of Rome Tor Vergata, Rome, Italy}
\affil[2]{A.A. Martinos Center for Biomedical Imaging and Harvard Medical School, Boston, USA}

\corr{tommaso.boccato@uniroma2.it}{TB}

\newcommand{\hptonedef}{\textbf{hypothesis 1 - h1.}}
\newcommand{\hptone}{\textbf{h1.}}
\newcommand{\hpttwo}{\textbf{h2.}}
\newcommand{\hptthree}{\textbf{h3.}}
\newcommand{\model}{SynaptoGen}
\newtheorem{proposition}{Proposition}

\newcommand{\biorules}{ which includes the genetic rules derived from \textit{C. elegans}}
\newcommand{\resfigcaption}[2][]{Mean reward distributions from the model families, characterized by a #2-gene profile#1, tested on the four selected RL environments. The green crosses represent the mean rewards, averaged over 10 episodes, achieved by the trained \model{} models. Each \textit{scatterplot} point represents the mean reward obtained by a specific agent. The black crosses denote instead the distribution means. Model families are color-coded. The dashed horizontal lines indicate the reward threshold beyond which the task associated with an environment is considered solved.}
\newcommand{\restabcaption}[2][]{Summary metrics computed over the reward distributions obtained from the model families characterized by a #2-gene profile#1. Three metrics are reported: the distribution mean, a mean computed over the top-10 agents, and the percentage of simulated agents that solved the task. Each group of rows refers to one of the selected RL environments and the best scores are highlighted in bold.}

\begin{document}

\maketitle

\begin{abstract}
There is growing consensus among neuroscientists that neural circuits critical for survival are the result of genomic decompression processes. We introduce \model{}, a novel computational framework--member of the Connectome Models family--bringing synthetic biological intelligence closer, facilitating neural biological agent development through precise genetic control of synaptogenesis. \model{} is the first model of its kind offering mechanistic explanation of synaptic multiplicity based on genetic expression and protein interaction probabilities. The framework connects genetic factors through a differentiable function, working as a neural network where synaptic weights equal average numbers of synapses between neurons, multiplied by conductance, derived from genetic profiles. Differentiability enables gradient-based optimization, allowing generation of genetic expression patterns producing pre-wired biological agents for specific tasks. Validation in simulated synaptogenesis scenarios shows agents successfully solving four reinforcement learning benchmarks, consistently surpassing control baselines. Despite gaps in biological realism requiring mitigation, this framework has potential to accelerate synthetic biological intelligence research.
\end{abstract}

\section{Introduction}

Let us consider a scenario in which, during brain development, \hptonedef{} \textit{it becomes possible to manipulate, prior to the formation of synapses, the gene expression profiles of individual neurons}. Furthermore, let us assume that \hpttwo{} \textit{we possess the capability to influence these expression profiles in a manner that directs synaptogenesis towards a specific neuronal network topology}. In this context, \hptthree{} \textit{we might be able to obtain the optimal computational graph, expressed as a composition of functions that represent the behaviour of neurons, required to solve a task of interest}. Small living organisms, or organoids \citep{BHADURI2020361}, could be, in principle, genetically programmed to develop into neuronal networks capable of solving pre-specified tasks. Such technology would lead to disruptive applications--e.g., extreme low-power computing, micro-devices for the control of biological systems, or novel biological testing platforms capable of accelerating drug discovery. To date, hypothesis \hptone{} seems to be verified, \citep{nishikawa_optimization_2014} while in the case of \hptthree{} we can partially rely on artificial neural networks and optimization techniques \citep{DBLP:journals/corr/KingmaB14,graves2014generating}.

\begin{figure}[ht!]
    \centering
    \includegraphics[width=\textwidth]{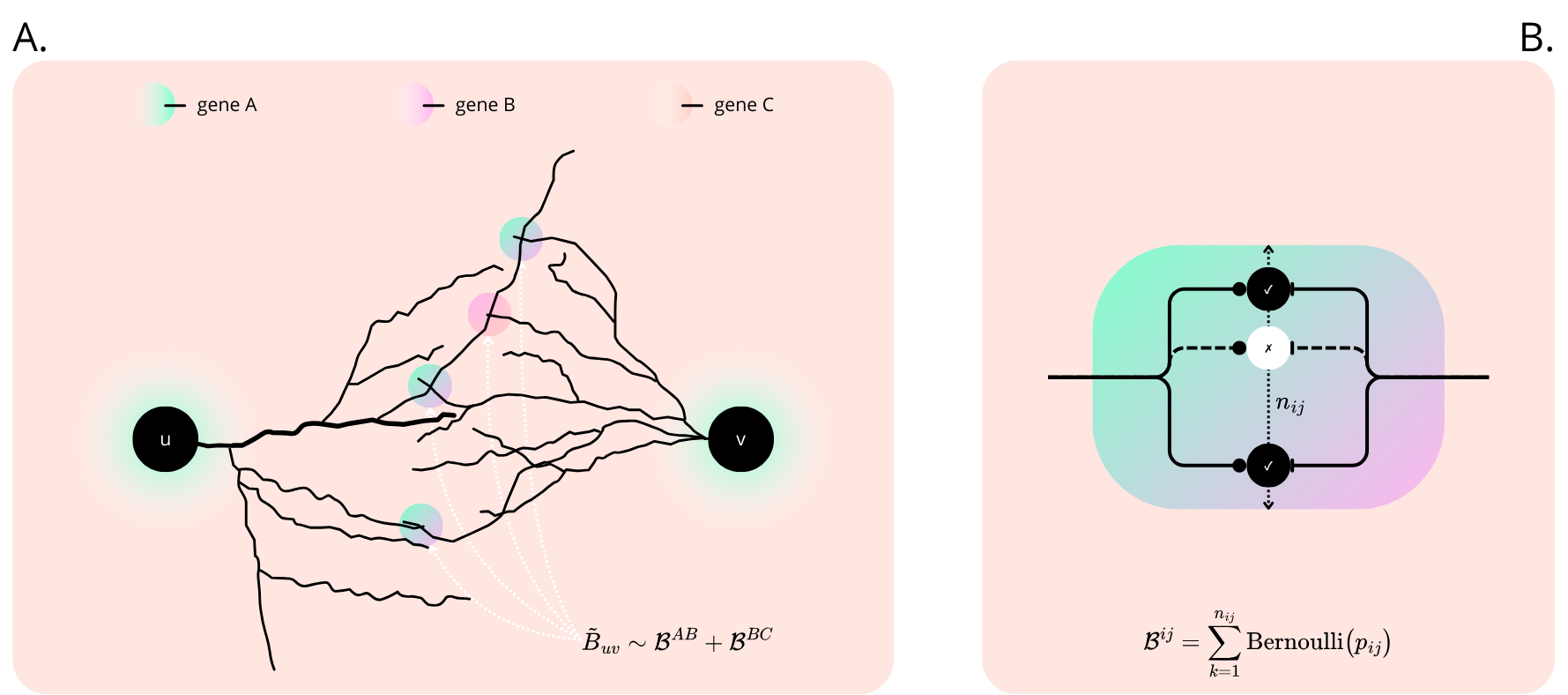}
    
    \vspace{.025\textwidth}
    
    \includegraphics[width=\textwidth]{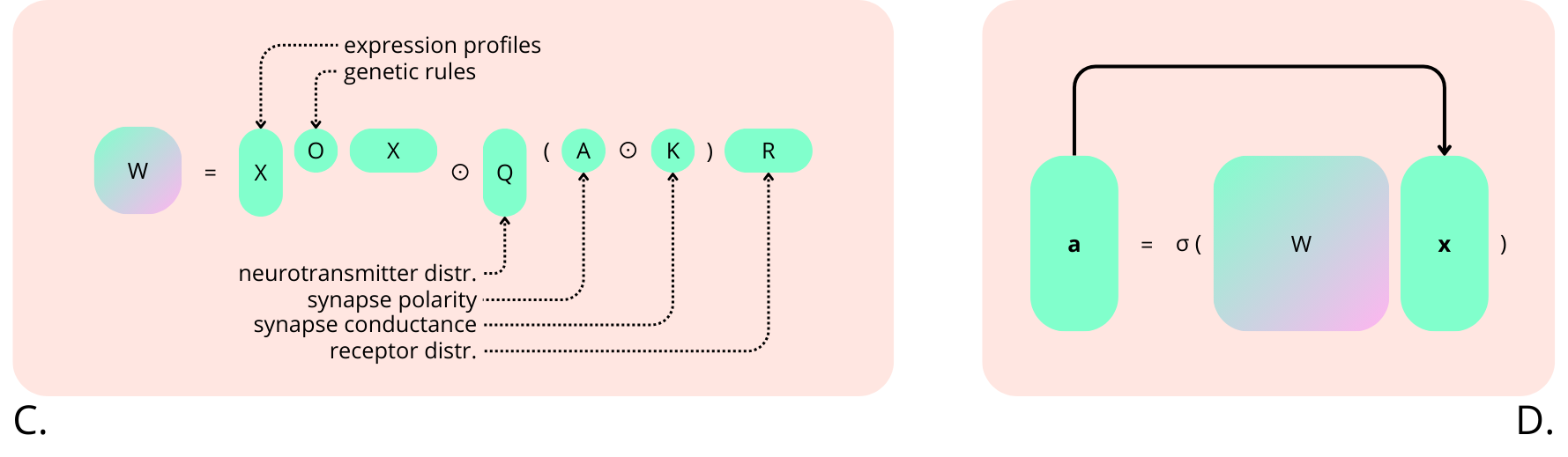}
    \caption{Multi-scale overview of the \model{} model. \textbf{A.} In neural networks, synaptogenesis--the formation of synapses--can be approximated as the outcome of interactions (e.g., molecular binding) between proteins translated from gene pairs. \model{} models this process as the realization $\tilde{B}_{uv}$ of a random variable defined as the sum of multiple binomial random variables (e.g., $\mathcal{B}^{AB} + \mathcal{B}^{BC}$), one for each potentially interacting gene pair. In the notation used, the $uv$ subscript indicates the neuronal pair taken into consideration. \textbf{B.} The use of binomial random variables stems from the idea of modeling synapse formation for each gene pair, as a process akin to flipping $n_{ij}$ biased coins, each representing a $\text{Bernoulli}(p_{ij})$ random variable. \textbf{C.} When working with matrix representations of the genetic factors in the panel, it is mathematically provable (see Section \nameref{sec:methods}) that a series of matrix multiplications and point-wise operations yields a matrix where the entry in the $u$-th row and $v$-th column contains the expected number of synapses formed between neurons $u$ and $v$, multiplied by their average synaptic conductance. \textbf{D.} This resulting matrix, $W$ ($\bar{W}$ in Section \nameref{sec:methods}), can be interpreted as a weight matrix and integrated into architectures such as multilayer perceptrons (MLPs). Here, $\bm{x}$ denotes an MLP layer's input, while $\bm{a}$ the resulting activations.}
    \label{fig:overview}
\end{figure}

In this work, we take a step toward the realization of the joint technology conceptualized in \hpttwo{} and \hptthree{} by proposing \model{}\footnote{\url{https://github.com/BoCtrl-C/synaptogen}} (Figure \ref{fig:overview}), a model that links, by means of differentiable functions, vector representations of gene expression profiles (i.e., the pattern of genes actively being transcribed into RNA in a cell) and genetic rules (i.e., interaction probabilities of protein pairs involved in synaptogenesis) to the average number of synaptic connections between pairs of neurons as well as their synaptic conductance values. We substantiate our work through theoretical development, which hinges on novel propositions and related mathematical proofs. \model{} is compatible with backpropagation and can be inserted in learning frameworks where optimization is performed through gradient descent, enabling management of network sizes and task complexities beyond the capabilities of other optimization techniques. Finally, \model{} is designed with flexibility in mind, allowing practitioners to choose which biological quantities to optimize (e.g., genetic rules, expression profiles, or both).

\subsection{Related Work}

Interest in leveraging organoid production for computational purposes has surged since 2022, following the introduction of DishBrain \citep{KAGAN20223952} by Cortical Labs. DishBrain combines in-vitro neural networks, cultured from human or rodent sources, with a simulated environment--specifically the game "Pong"--via a high-density multielectrode array. The system employs the \textit{free energy principle}, which suggests that neural networks adapt by minimizing the unpredictability of their sensory inputs through belief updating and environmental interactions. Although the neuron cultures in DishBrain exhibited statistically significant improvements in gameplay performance (e.g., \textit{average rally length}, \textit{\% of aces} and \textit{\% of long-rallies}) compared to the control conditions defined by the authors (i.e., random agents), it is difficult to conclude that the neuronal networks fully mastered the task.

In contrast to DishBrain, our research aims to ``train'' networks of neurons by influencing synaptogenesis through genetic manipulations at the individual neuron level, a largely unexplored task in the literature, with only a limited number of closely related studies available. The basis for our work is outlined in \citep{barabasi_constructing_2021,BARABASI2020435}, which introduces methods for constructing networks based on genetic encodings inspired by the wiring rules of the brain. These methods were further elaborated in the Connectome Model (CM) \citep{doi:10.1073/pnas.2009093117}, where the authors decomposed the adjacency matrix of a connectome into the product of three matrices representing specific genetic quantities. Another development was presented in \citep{barabasi_complex_2023}, where the CM's matrix entries were treated as learnable parameters, resulting in the weight matrix of a Multilayer Perceptron (MLP) within the context of training neural networks. While this methodology has proven effective in producing parameter-efficient neural networks, it maintains a notable distance from the biological intricacies of real neuronal networks. A distinct generalization of the CM has also been proposed for the computational inference of synaptic polarities \citep{https://doi.org/10.1002/advs.202104906}, a quantity not considered in the original CM.

Similarly, our work draws inspiration from the CM but is geared towards a more bio-plausible computational modeling of synaptogenesis, with the novel elements extensively discussed in Section \nameref{sec:methods}.

\section{Methods}\label{sec:methods}

In 2020, Kovács et al. proposed the CM, a novel  strategy to link a brain’s connectome ($B$) to the expression patterns of individual neurons ($X$) and existing biological mechanisms-- or genetic rules--$O$:
\begin{equation}\label{eq:cm}
    B = XOX^T
\end{equation}
In the CM's first interpretation, each row of $X$ referred to a specific neuron while the $i$-th entry of the row described the binary expression (1--``expressed''--or 0--``not expressed'') of gene $i$, one of the genes involved in synapse formation. Matrix $O$, instead, represented interaction compatibility, in the synapse formation process, for proteins translated from all gene pairs. Hence, $X \in \{0, 1\}^{N \times G}$ and $O \in \{0, 1\}^{G \times G}$ were defined as binary matrices while the entries of $B$, of shape $N \times N$, belonged to $\mathbb{Z^+}$; with $N$ and $G$ denoting the number of neurons and genes, respectively. When $G \ll N$ however, a very common scenario in nature \citep{Koulakov2021.03.16.435261}, not all possible connectomes can be decomposed through \eqref{eq:cm}. For this reason, the  authors of the CM went on to relaxing the genetic rules matrix to $O \in [0, 1]^{G \times G}$, interpreting its entries as probabilities, which generates the following approximation:
\begin{align}
    B &\simeq XOX^T,\\
    O &= \arg\min_{O'} ||B - XO'X^T||^2
\end{align}
where $||\bullet||$ is intended as the Frobenius norm.

In this paper, we formulate a more general alternative to this framework by building a model that takes into account synaptic multiplicity and conductance. Starting from \eqref{eq:cm}, we design two novel interpretations tightly linked to the formalism with which the quantities of interest (i.e., the number of synapses between neurons and their conductances) have been represented in our model. Our theoretical framework is as follows.

Let the number of synaptic connections between two neurons be represented by the following:
\begin{equation}
   \mathcal{B} = \sum_{i,j} \mathcal{B}^{ij} 
\end{equation}
where $\mathcal{B}^{ij}$ is a binomial random variable that expresses the contribution of the $(i, j)$ gene pair to the total synaptic count:
\begin{equation}
    \mathcal{B}^{ij} = \text{Bin}(n_{ij}, p_{ij})
\end{equation}
And let $\bm{x} \in {\mathbb{R}^+}^G$ and $\bm{y} \in {\mathbb{R}^+}^G$ be vector representations of gene expression in the pre- and post-synaptic neurons, respectively.
\begin{proposition}\label{th:number-synapses}
    If the product between the $i$-th entry of $\bm{x}$ and the $j$-th entry of $\bm{y}$ denotes the number of independent experiments that characterizes $\mathcal{B}^{ij}$--i.e., $x_i y_j = n_{ij}$--and entry $O_{ij}$ corresponds to probability $p_{ij}$, then the expected number of synapses between two neurons can be calculated as:
    \begin{equation}\label{eq:avg-num-syn}
        \mathbb{E}[\mathcal{B}] = \bm{x}^TO\bm{y}
    \end{equation}
\end{proposition}
\begin{proof}
    From probability theory,
    $$\mathbb{E}[\mathcal{B}^{ij}] = n_{ij}p_{ij}$$
    and due to the linearity of expectation, we have
    $$\mathbb{E}[\mathcal{B}] = \sum_{i,j} n_{ij}p_{ij}$$
    On the other hand,
    \begin{align*}
    \bm{x}^TO\bm{y} &= \bm{x}^T[\dots, \sum_j y_j O_{ij}, \dots]^T\\
    &= \sum_i x_i \sum_j y_j O_{ij}\\
    &= \sum_{i,j} x_i y_j O_{ij}
    \end{align*}
    Recalling that $x_i y_j = n_{ij}$ and $O_{ij} = p_{ij}$, the proof is concluded.
\end{proof}
In different terms, if the hypotheses of Proposition \ref{th:number-synapses} are verified, gene expression in a pair of genes tells us how many attempts we can make to place a synapse between a pre- and a post-synaptic neuron; the genetic rule, instead, describes the probability of success, conditioned on the interaction between the proteins translated from the corresponding genes, for  each attempt. It is worth noting that \eqref{eq:avg-num-syn} represents the average number of links between two specific nodes of the connectome. Keeping in mind that genetic rules are shared across neurons, the equation can easily be generalized to the whole connectome:
\begin{equation}\label{eq:avg-connectome}
    \bar{B} = \mathbb{E}[B] = XOX^T
\end{equation}
where $X$ is obtained by stacking the expression profiles of all neurons (e.g., $X^T = [\dots, \bm{x}, \dots, \bm{y}, \dots]$).

In order to model synaptic conductances, a slightly more complex formalism is required. We restrict ourselves to chemical synapses, which are the result of the interplay between neurotransmitters released by pre-synaptic neurons and receptors in post-synaptic neurons. We also account for the fact that a chemical synapse can also have an excitatory or inhibitory effect depending on the nature of the receptor that receives a specific neurotransmitter \citep{10.1371/journal.pcbi.1007974,https://doi.org/10.1002/advs.202104906}. The way in which synapses work in our framework is described by the following equation:
\begin{equation}
    I_v = \sum_uG_{uv}V_u
\end{equation}
where $I_v$ is the current injected into post-synaptic neuron $v$ while $V_u$ is an input voltage from pre-synaptic neuron $u$; $G_{uv}$, is the equivalent conductance that takes care of all the synapses formed between $u$ and $v$. To model the possibility of characterizing synapses by different neurotransmitter-receptor pairs, we rely again on random variables as follows:

Let $\mathcal{T}$ be a multinomial random variable representing the process of randomly picking, from $u$, a synaptic vesicle filled with a specific neurotransmitter. And let $\mathcal{R}$ be a multinomial random variable representing the process of randomly selecting a specific receptor from the membrane of $v$. We define vectors $\bm{q} \in [0, 1]^L$ and $\bm{r} \in [0, 1]^M$ as the probability distributions associated with $\mathcal{T}$ and $\mathcal{R}$; where $L$ denotes the total number of neurotransmitters while $M$ the number of receptors. We also define $A \in \{-1, 0, 1\}^{L \times M}$ as the polarity matrix (for further details, refer to Appendix \nameref{sec:polarity}) and $K \in {\mathbb{R}^+}^{L \times M}$ as the conductance matrix. In detail, entry $A_{ij}$ defines the polarity of synapses derived from the interaction of the $i$-th neurotransmitter with the $j$-th receptor ($A_{ij} = 0$ if the considered neurotransmitter and receptor are not compatible) while $K_{ij}$ stores its linked conductance. We finally set $\mathcal{G} = f(\mathcal{T}, \mathcal{R})$, with $f(i, j) = A_{ij}K_{ij}$. In other words, $\mathcal{G}$ represents the ``signed'' conductance of a synapse randomly selected from the ones that connect neurons $u$ and $v$.
\begin{proposition}\label{th:expected-conductance}
    If $\mathcal{T}$ and $\mathcal{R}$ are independent (i.e., the distribution of receptors in the post-synaptic neuron does not depend on the neurotransmitters synthesized by the pre-synaptic neuron), the expected ``signed'' conductance of a randomly picked synapse can be calculated as:
    \begin{equation}\label{eq:avg-conductance}
        \mathbb{E}[\mathcal{G}] = \bm{q}^T (A \odot K) \bm{r}
    \end{equation}
\end{proposition}
\begin{proof}
    By expanding the matrix multiplications in \eqref{eq:avg-conductance}, we have:
    \begin{align*}
        \bm{q}^T (A \odot K) \bm{r} &= \bm{q}^T[\dots, \sum_j r_j A_{ij} K_{ij}, \dots]^T\\
        &= \sum_i q_i \sum_j r_j A_{ij} K_{ij}\\
        &= \sum_{i,j} q_i r_j A_{ij} K_{ij}
    \end{align*}
    Thanks to the independence of $\mathcal{T}$ and $\mathcal{R}$:
    $$\mathbb{P}[\mathcal{T} = i, \mathcal{R} = j] = q_i r_j$$
    where $\mathbb{P}[\bullet]$ stands for ``probability of''. Hence,
    $$\bm{q}^T (A \odot K) \bm{r} = \sum_{i,j} \mathbb{P}[\mathcal{T} = i, \mathcal{R} = j] f(i, j)$$
    that corresponds exactly to the definition of $\mathbb{E}[\mathcal{G}]$.
\end{proof}
As for \eqref{eq:avg-connectome}, also \eqref{eq:avg-conductance} can be generalized by stacking the neurotransmitter distributions of all pre-synaptic neurons in $Q = [\dots, \bm{q}, \dots]^T$ and the receptor distributions of post-synaptic neurons in $R = [\dots, \bm{r}, \dots]^T$:
\begin{equation}\label{eq:avg-conductance-connectome}
    \bar{G} = \mathbb{E}[G] = Q(A \odot K)R^T
\end{equation}

As a next step, \eqref{eq:avg-connectome} and \eqref{eq:avg-conductance-connectome} can be inserted into the core equation of our model, which follows:
\begin{align}\label{eq:model}
    \bar{W} &= \bar{B} \odot \bar{G}\\
    &= (XOX^T) \odot (Q(A \odot K)R^T)\nonumber
\end{align}
Summarizing, through \eqref{eq:model} we are able to express the average equivalent conductance between all pairs of neurons as a differentiable function of their gene expression profiles and distributions of synthesized neurotransmitters and receptors, which, in turn, depend on gene expression. Furthermore, in our formalism, synaptogenesis can be simulated by sampling from the random variables we have defined. For instance, the simplest approximation of synaptogenesis can be obtained as follows (further details in Appendices \nameref{sec:simulation}, \nameref{sec:quantization}):
\begin{equation}
    W = \tilde{B} \odot \bar{G}
\end{equation}
with
\begin{equation}
    \tilde{B} \sim
    B = \begin{bmatrix}
        \dots & \dots & \dots\\
        \dots & \mathcal{B}_{uv} & \dots\\
        \dots & \dots & \dots
    \end{bmatrix}
\end{equation}
where $\sim$ stands for ``sampled from''.

\section{Results}\label{sec:results}

To validate the proposed framework and assess its applicability in real-world scenarios, we conducted a series of experiments that involved simulating synaptogenesis in small populations of neurons. In simple terms, we simulated the formation of synapses in neuron populations where genes and gene expression were manipulated at the level of individual neurons. These manipulations followed the genetic rules and expression profiles optimized by our model with the aim of enabling the resulting fully-developed neuronal networks to perform effectively in a pre-specified task.

Our approach begins with a simplified setup, where a neuronal population comprises spatially-separated layers. We assumed that neurons in one layer could only attempt to form synapses with neurons in adjacent layers. This restriction allowed us to focus on multipartite network topologies and implement \model{} as a customized MLP, where its weight matrices are decomposed according to \eqref{eq:model}, with genetic rules shared across layers.

Regarding the task for our proof-of-concept experiments, we chose reinforcement learning (RL) as a bio-plausible benchmark. Biological organisms equipped with neuronal networks, indeed, exist immersed in an external environment from which they receive stimuli and upon which they can perform actions. These same characteristics are also inherent to the tasks addressed by RL. Our goal was to create virtual neural agents capable of solving the control tasks defined by the \texttt{CartPole-v1}, \texttt{MountainCar-v0}, \texttt{LunarLander-v2} and \texttt{Acrobot-v1} environments from the OpenAI Gym library \citep{1606.01540}, among the most famous benchmarks in the RL community. In Cart Pole, a pole is attached to a cart that moves along a frictionless track. The objective is to balance the pole by applying forces to move the cart left or right. In Mountain Car, instead, the goal is to strategically accelerate a car to climb a sinusoidal valley-and-hill terrain. In Lunar Lander, agents must operate the main and orientation engines of a lander to ensure it lands smoothly on a designated landing pad without damage. Finally, in the Acrobot game, the aim is to apply torques to the actuated joint of a two-link chain to make the free end of the chain surpass a given height.

\begin{sloppypar}
Initially, we separately trained \model{}--16/32/64 genes, 3 neurotransmitters, neuronal population sizes of ${\sim}128$ (see Appendix \nameref{sec:train-details} for the training details)--on the four selected environments using the DQN algorithm \citep{mnih_human-level_2015}. The training provided the learned genetic rules, gene expression profiles, neurotransmitter and receptor distributions, and synaptic conductances that enabled the agents to perform the tasks. Notably, \model{} is built on the average equivalent conductances introduced in \eqref{eq:avg-conductance-connectome}, and can be interpreted as an average agent that reflects the effects of the underlying genetically-derived quantities.
\end{sloppypar}

Successively, for each environment, we sampled 100 neural networks from the trained \model{} models, simulating the development, based on the computed synaptogenesis rules, of multiple populations of neurons into neuronal networks. We then measured their performance and compared the obtained metrics with those from two carefully designed baselines. The first baseline (Appendix \nameref{sec:snes}) adopts the optimization technique employed in \citep{Stockl2021.05.18.444689}, which has been applied to optimize related models, the \textit{probabilistic skeletons}, for similar tasks, i.e., the optimization of connection probabilities between neuronal types. This technique, known as SNES \citep{10.1145/2330784.2330815}, is an evolutionary algorithm specifically designed for medium-to-large problem dimensions. Given the structure of \model{}, which is explicitly designed for gradient-based optimization but remains general and flexible enough to accommodate other  optimizers, SNES could serve as both a benchmark we aim to surpass and a potential alternative worth exploring. The second baseline (Appendix \nameref{sec:bio-plausible-init}) leveraged the \model{}'s biology, using the optimized genetic rules $O$ and conductance matrix $K$, while initializing gene expression profiles in a bio-plausible manner following the procedure identified by Kerstjens et al. \citep{kerstjens_constructive_2022,Kerstjens2024.09.19.613507}. As before,  100 neuronal networks were sampled from each baseline.

\begin{figure}[h!]
    \centering
    \includegraphics[width=\textwidth]{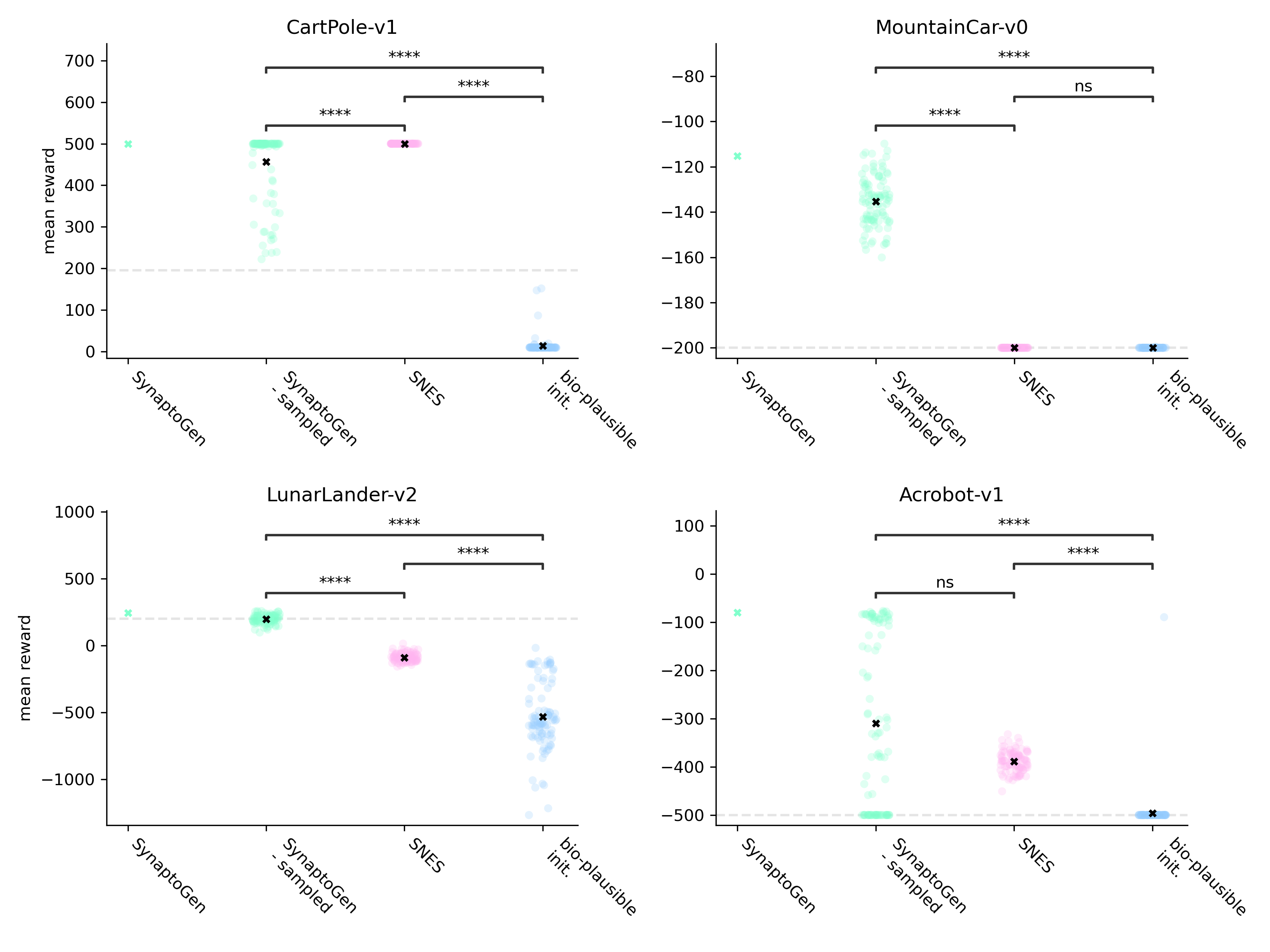}
    \caption{\resfigcaption{16}}
    \label{fig:reward-dist-16}
\end{figure}

\begin{table}[h!]
    \centering

    \begin{tabular}{llrrrrr}
    \toprule
     &  & \textbf{mean} & \textbf{std} & \textbf{top-10 mean} & \textbf{top-10 std} & \textbf{\% solved} \\
    \textbf{environment} & \textbf{model} &  &  &  &  &  \\
    \midrule
    \multirow[c]{3}{*}{CartPole-v1} & SynaptoGen & 456.05 & 83.38 & \textbf{500.00} & 0.00 & \textbf{100.00\%} \\
     & SNES & \textbf{500.00} & 0.00 & \textbf{500.00} & 0.00 & \textbf{100.00\%} \\
     & bio-plausible & 13.56 & 21.12 & 51.12 & 56.24 & 0.00\% \\
    \cmidrule{2-7}
    \multirow[c]{3}{*}{MountainCar-v0} & SynaptoGen & \textbf{-135.26} & 11.17 & \textbf{-115.75} & 3.31 & \textbf{100.00\%} \\
     & SNES & -200.00 & 0.00 & -200.00 & 0.00 & 0.00\% \\
     & bio-plausible & -200.00 & 0.05 & -199.95 & 0.16 & 1.00\% \\
    \cmidrule{2-7}
    \multirow[c]{3}{*}{LunarLander-v2} & SynaptoGen & \textbf{197.92} & 32.37 & \textbf{250.12} & 7.60 & \textbf{47.00\%} \\
     & SNES & -88.92 & 31.67 & -28.71 & 17.71 & 0.00\% \\
     & bio-plausible & -530.68 & 255.51 & -115.53 & 36.15 & 0.00\% \\
    \cmidrule{2-7}
    \multirow[c]{3}{*}{Acrobot-v1} & SynaptoGen & \textbf{-309.37} & 179.32 & \textbf{-80.38} & 2.39 & 63.00\% \\
     & SNES & -388.44 & 21.69 & -350.77 & 9.95 & \textbf{100.00\%} \\
     & bio-plausible & -495.89 & 41.07 & -458.93 & 129.87 & 1.00\% \\
    \bottomrule
    \end{tabular}

    \caption{\restabcaption{16}}
    \label{tab:aggregated-16}
\end{table}

The results obtained from experiments with agents characterized by a genetic profile of 16 genes are shown in Figure \ref{fig:reward-dist-16}. Table \ref{tab:aggregated-16} provides summary metrics of the outcomes of each experiment. The first metric is the average reward calculated across groups of 100 sampled agents. The second metric is also the average reward, but calculated for the top-10 best-performing agents in each group. This metric aims to assess the models' effectiveness in scenarios where the entity responsible for synthesizing biological agents prioritizes achieving a small subset of highly performing agents, even at the cost of the overall process yield--the number of agents capable of solving their task divided by the total number of agents. Finally, the third metric describes the percentage of simulated agents capable of solving the tasks defined by the four selected environments (see Appendix \nameref{sec:resolution} for further discussion). This metric represents the global yield of each methodology under consideration.

In terms of average reward \model{} (trained via gradient descent) emerged as the best-performing model, outperforming agents optimized with SNES in three out of four environments. It also consistently surpassed the baseline characterized by biologically plausible initialization of gene expression. The SNES agents also performed well, exceeding the bio-plausible baseline--the one characterized by the bio-plausible expression initialization--in three out of four environments. When examining the average rewards of the top-10 agents, the superiority of \model{} and its derived agents becomes even more evident. They achieved the highest performance across all environments studied, sharing the leading position only in Cart Pole. By contrast, the SNES top-10 agents achieved average rewards indicative of task resolution in only 50\% of the cases. For the bio-plausible baseline, nearly all top-10 agents failed to achieve sufficient performance. Finally, regarding the percentage of agents capable of solving the tasks their genetic profiles were optimized/initialized for, \model{} was the only approach to achieve non-zero percentages in all four environments. Moreover, it achieved the highest percentage in three out of four cases. SNES agents also performed reasonably well, reaching a 100\% resolution rate in half the environments, though failing entirely in the other half. For the bio-plausible baseline, only 2 out of 400 sampled agents successfully solved their respective tasks. Notably, in 10 out of 12 comparisons, the differences between reward distributions were statistically significant based on the Mann-Whitney test with Bonferroni correction (for the multiple environment-specific comparisons performed). For brevity and clarity, results for agents derived from genetic profiles with 32 or 64 genes are presented in Appendix \nameref{sec:additional-results}. These results exhibit minimal differences from those reported here and lead to equivalent conclusions.

\begin{figure}[h!]
    \centering
    \includegraphics[width=\textwidth]{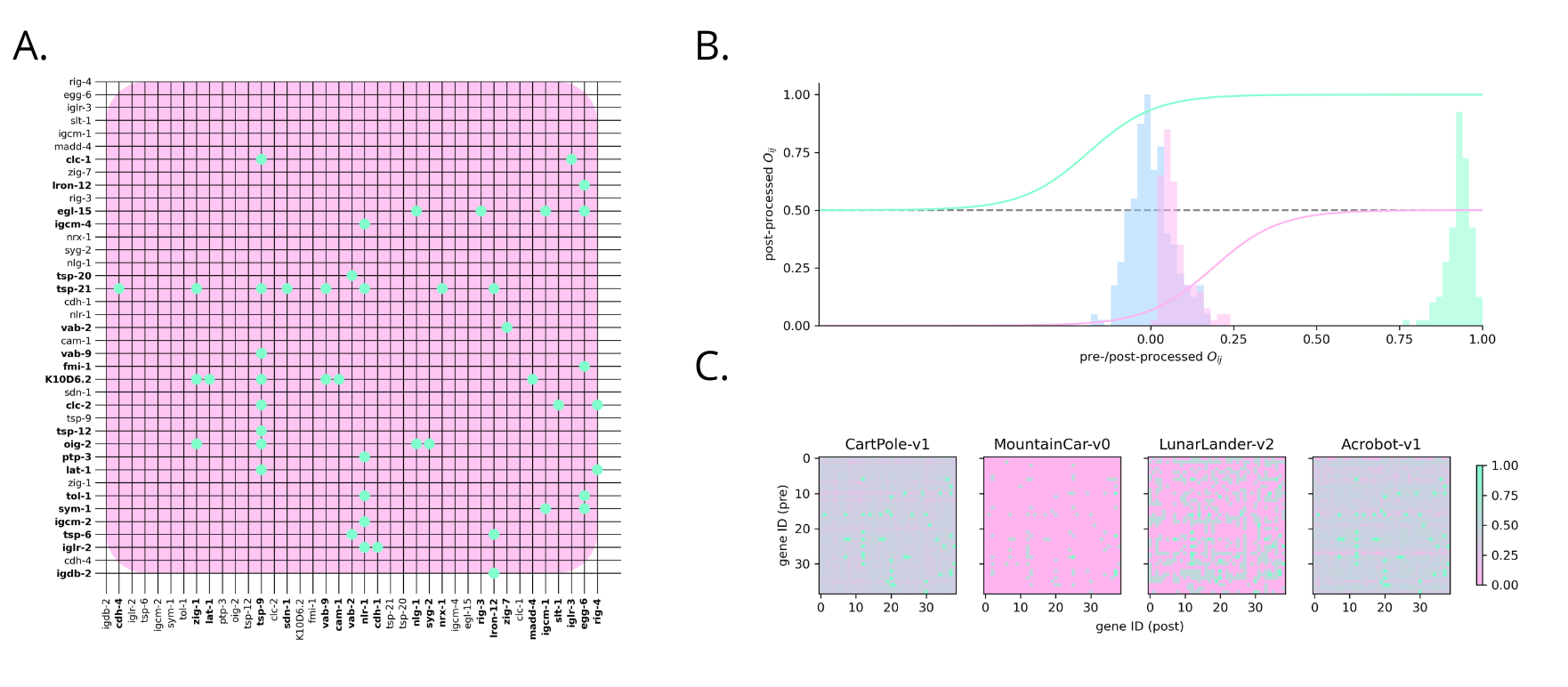}
    \caption{\textbf{A.} Co-expression data computed following our extended nDGE variant from the expression patterns released with \citep{TAYLOR20214329}. In the matrix, rows correspond to genes expressed in pre-synaptic neurons, while columns represent genes utilized in post-synaptic neurons. The green circles indicate pairs of genes that are co-expressed in neurons that are connected but not co-expressed in neurons that could be connected but lack synapses. Genes involved in co-expressed pairs are highlighted in bold. \textbf{B.} Visualization of the two sigmoids used to map the learnable parameters associated with the genetic rules into the probabilities in $O$. The green sigmoid is applied to the parameters corresponding to co-expressed pairs, while the pink sigmoid is applied to the remaining ones. We also show in blue the parameter distribution after initialization and in green (co-expressed pairs) and pink (non-co-expressed pairs) the distributions of the probabilities obtainable in an example scenario where 50\% of the gene pairs are co-expressed. \textbf{C.} Genetic rules learned by \model{} in our bio-plausible validations. The emerging patterns are fully consistent with the co-expression data in A. (correlation coefficients: 0.92, 0.74, 0.44, 0.84; same order as in C.) and demonstrate how the model assigned high probabilities (${\sim}1$) specifically to the genetic rules corresponding to the pairs of co-expressed genes in the \textit{C. elegans} nerve ring.}
    \label{fig:bio-rules}
\end{figure}

\begin{figure}[h!]
    \centering
    \includegraphics[width=\textwidth]{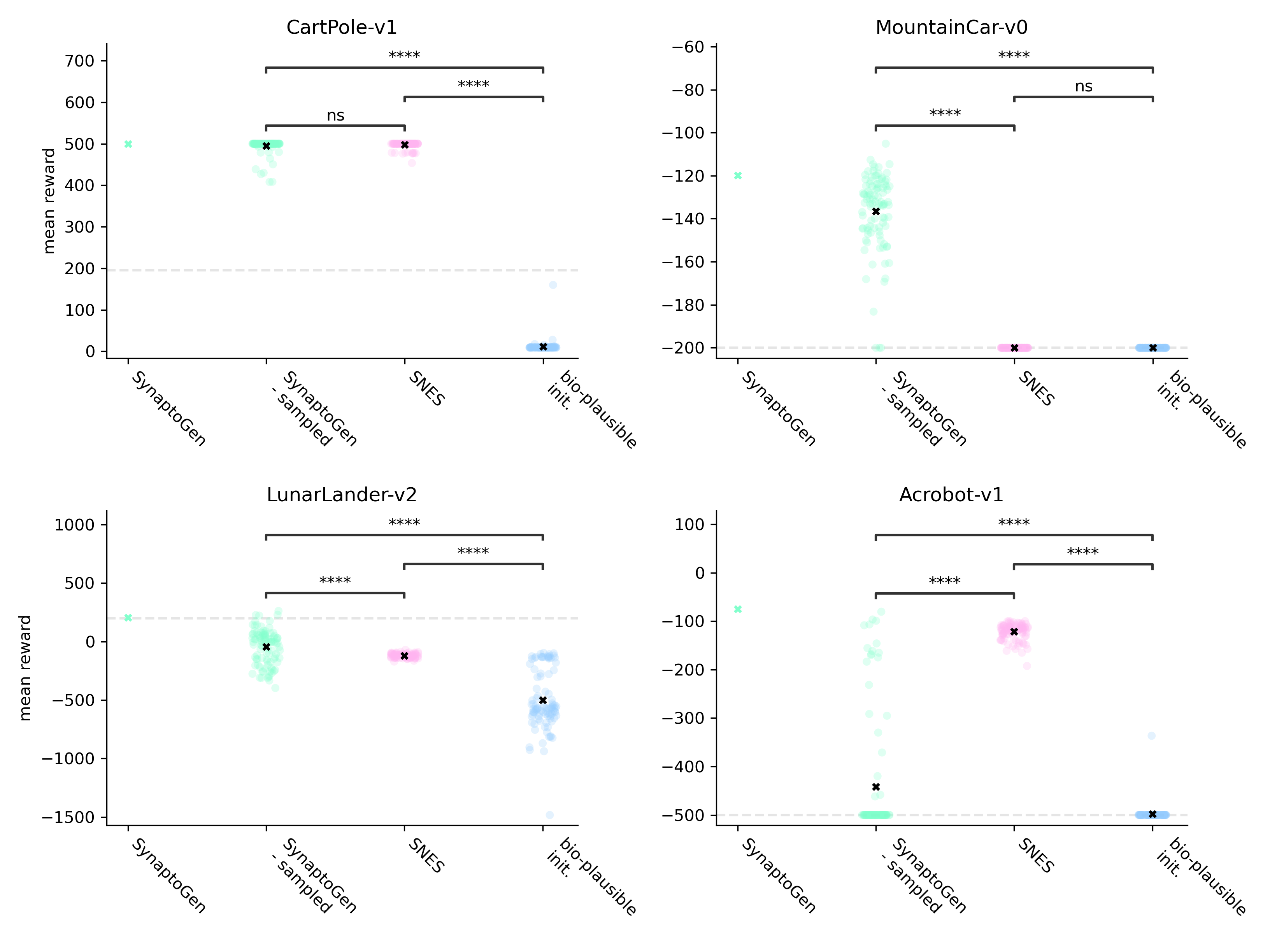}
    \caption{\resfigcaption[\biorules]{39}}
    \label{fig:reward-dist-bio}
\end{figure}

\begin{table}[h!]
    \centering

    \begin{tabular}{llrrrrr}
    \toprule
     &  & \textbf{mean} & \textbf{std} & \textbf{top-10 mean} & \textbf{top-10 std} & \textbf{\% solved} \\
    \textbf{environment} & \textbf{model} &  &  &  &  &  \\
    \midrule
    \multirow[c]{3}{*}{CartPole-v1} & SynaptoGen & 494.39 & 18.27 & \textbf{500.00} & 0.00 & \textbf{100.00\%} \\
     & SNES & \textbf{497.25} & 8.09 & \textbf{500.00} & 0.00 & \textbf{100.00\%} \\
     & bio-plausible & 11.26 & 15.16 & 28.44 & 46.52 & 0.00\% \\
    \cmidrule{2-7}
    \multirow[c]{3}{*}{MountainCar-v0} & SynaptoGen & \textbf{-136.53} & 18.06 & \textbf{-115.21} & 4.08 & \textbf{97.00\%} \\
     & SNES & -200.00 & 0.00 & -200.00 & 0.00 & 0.00\% \\
     & bio-plausible & -200.00 & 0.00 & -200.00 & 0.00 & 0.00\% \\
    \cmidrule{2-7}
    \multirow[c]{3}{*}{LunarLander-v2} & SynaptoGen & \textbf{-45.59} & 146.52 & \textbf{184.01} & 47.31 & \textbf{4.00\%} \\
     & SNES & -120.70 & 16.98 & -91.96 & 7.96 & 0.00\% \\
     & bio-plausible & -499.67 & 253.46 & -116.52 & 12.17 & 0.00\% \\
    \cmidrule{2-7}
    \multirow[c]{3}{*}{Acrobot-v1} & SynaptoGen & -441.75 & 126.19 & -128.74 & 33.49 & 21.00\% \\
     & SNES & \textbf{-121.59} & 16.50 & \textbf{-102.17} & 1.84 & \textbf{100.00\%} \\
     & bio-plausible & -498.37 & 16.34 & -483.66 & 51.67 & 1.00\% \\
    \bottomrule
    \end{tabular}

    \caption{\restabcaption[\biorules]{39}}
    \label{tab:aggregated-bio}
\end{table}

To provide additional evidence of \model{}’s effectiveness, we designed an additional set of simulations aimed at validating the framework in a context closer to the biological reality of neuronal networks. In these experiments, we used  genetic rules derived from experimental data collected on the nerve ring of the nematode \textit{C. elegans}. This region was selected due to the availability of both transcriptomic data \citep{TAYLOR20214329} and connectomic data--including membrane contacts and synapses \citep{brittin_multi-scale_2021,cook_whole-animal_2019,witvliet_connectomes_2020}--collected from it. The genetic rules were constructed based on a restricted set of 141 cell adhesion molecules (CAMs) with a documented role in synapse formation \citep{doi:10.1126/science.1143762,10.7554/eLife.29257}, following a procedure adapted from the Network Differential Gene Expression Analysis (nDGE) framework proposed in \citep{TAYLOR20214329}. The procedure is described in Appendix \nameref{sec:bio-rules} and summarized in Figure \ref{fig:bio-rules}. Briefly, we first identified pairs of genes responsible for CAM production that were co-expressed in nerve ring neurons with synaptic connections but not co-expressed in cells sufficiently close to form synapses yet lacking connections. We then constrained the (learnable) genetic linked to these pairs to assume high probabilities ($> 0.5$), while assigning low probabilities ($< 0.5$) to all other rules.

The results of these experiments are presented in Figure \ref{fig:reward-dist-bio} and Table \ref{tab:aggregated-bio}. In this context, the SNES agents improved their competitiveness in terms of average reward, surpassing \model{} in Acrobot. Nevertheless, \model{} maintained the highest average reward in Mountain Car and Lunar Lander. As in the previous set of experiments, the bio-plausible baseline consistently ranked last across all environments. The average reward calculated for the top-10 agents followed a similar trend, with \model{} managing to achieve a tie in Cart Pole. Finally, in terms of resolution percentages, agents derived from \model{} once again were the only ones to achieve non-zero percentages across all tested environments. Results for SNES agents remained polarized, with all simulated agents resolving the tasks in two environments but none performing adequately in the remaining two. The bio-plausible baseline again failed to produce agents capable of solving any task.

\section{Discussion}

In this section, we will discuss the contributions of \model{} to neuroscience, with a particular focus on the ``genomic bottleneck'' field \citep{zador_critique_2019}, and its potential impact on technologies aimed at creating synthetic biological intelligence. We will also address the limitations of \model{} in its current form and propose strategies to mitigate these challenges.

Firstly, \model{} represents the first model capable of rigorously explaining synaptic multiplicity based on genes, their interaction probabilities, and gene expression, achieving (by design) unprecedented granularity in modeling synaptogenesis. Secondly, \model{} establishes a direct relationship between the average number of synapses between neurons, their average synaptic weights, and vector representations of genetic expression and protein interaction probabilities through a differentiable function. This is a key feature, as it ensures compatibility with gradient-based optimization techniques, bringing several advantages discussed further below. Notably, since \model{} can be deployed as a standard neural network, it broadens the accessibility of the simulation tools used in computational neuroscience to deep learning practitioners, whose contributions could significantly advance the field. Additionally, \model{} can be integrated with any differentiable AI model.

Arguably, one of the most significant features of \model{} is its performance. Simulations confirm that \model{} generates genetic profiles capable of guiding the development of biological agents that perform effectively, in terms of mean reward and yield, in the environments for which they have been pre-wired. Agents derived from \model{} consistently outperform the bio-plausible baseline across all experiments. This is not surprising, as populations of neurons left to interact without regulation are unlikely to form the connections necessary for task-specific behavior. More unexpectedly, \model{} also outperforms agents optimized through SNES in most scenarios. Unlike SNES, \model{} does not rely on evaluating the simulated agents' performance during optimization, highlighting the strategic advantage of gradient-based methods in our framework. We hypothesize that gradient descent scales better with problem size and task complexity. Supporting this hypothesis are results from the Mountain Car and Lunar Lander environments. In these environments, generally regarded by the RL community as more complex than inverted pendulum tasks, the performance gap between the \model{} agents and the SNES-derived ones was greater. Nonetheless, \model{} remains flexible and general enough to support alternative optimization techniques if advantageous. Notably, \model{} demonstrates strong effectiveness in ``biologically realistic'' simulations (i.e., simulations in which rules derived from co-expression data are utilized), maintaining its superiority despite the additional constraints introduced by biological genetic rules.

Despite these advances, there remains a considerable gap between our model and the biological reality of neuronal networks before it can be applied to real-world contexts. One major difference lies in the event-based nature of processing in biological neural networks. \model{} can be adapted into a spiking neural network without sacrificing differentiability by using libraries like \textit{snnTorch} \citep{eshraghian2021training}, which leverages surrogate gradients. This does not preclude exploring more complex neuron models, although such complexity may be unnecessary, as recent studies indicate that networks trained with simplified neuron models can be effectively deployed on neuromorphic hardware with negligible performance loss \citep{cakal_gradient-descent_2024}. This result indeed hints at the fact that even simplified neuron models could represent the essential computational dynamics of biological networks. Another significant assumption in our model is that synapses are modeled as simple resistors. Whether this level of abstraction suffices for developing functional biological networks remains an open question.

Moreover, the current implementation of \model{} assumes a feedforward multipartite network topology, which can be achieved by genetically inhibiting, through an additional set of genes, all synapses that could form between compatible neurons in non-adjacent layers. Another possiblity would be to employ an external pruning process. However, biological neuronal networks exhibit more complex, often cyclic wiring. \model{} can be extended to account for this complexity by integrating its weight matrix decomposition into frameworks like \textit{4Ward} \citep{BOCCATO2024127058,BOCCATO2024215}, which are designed to convert arbitrary graphs into neural networks trainable via backpropagation. Incorporating more complex topologies would also enable compliance with constraints derived from contactomic data (i.e., information about the physical contacts between neurons), as synapses can only form between neurons in close proximity. Using such topologies, \model{} could learn the average number of synapses only for neuron pairs which are actually in contact. Another phenomenon which his not modeled in the current version of \model{} is the modulation of genetic rules based on the distance between neurons. As highlighted in \citep{Stockl2021.05.18.444689}, connection probabilities decay exponentially with increasing distance. This behavior can be incorporated into \model{} without compromising differentiability by assigning each neuron a positional embedding  to calculate distances. This embedding could be learnable, given as a prior, or derived from genetic expression. We recall that a function such as $e^{-||\bm{p}^u - \bm{p}^v||}$, where $\bm{p}^u$ and $\bm{p}^v$ are the positions of neurons $u$ and $v$, is differentiable w.r.t. the positions except at coincident points.

Finally, the bio-plausibility of \model{} could be enhanced by integrating experimental data on the circuit-level properties of synapses formed by specific gene interactions--parameters  - these are currently treated as learnable. While the data available for integration is limited, promising progress by various research groups suggests this limitation may soon be alleviated \citep{TAYLOR20214329,doi:10.1073/pnas.2009093117,10.1371/journal.pcbi.1007974,https://doi.org/10.1002/advs.202104906}. Additionally, developing more faithful synaptogenesis simulation techniques--such as replacing average conductances with further sampling from the learned distributions--could make \model{} even more biologically accurate.

\subsection{Conclusions}

In this work, we introduced \model{}, a novel computational framework aimed at advancing the field of synthetic biological intelligence by addressing a critical challenge: controlling the genetic factors underlying synaptogenesis to develop neural agents pre-wired for task-specific behavior. By modeling synaptic multiplicity through gene expression and protein interaction probabilities, \model{} extends the existing frameworks by offering a granular approach to understanding neural connectivity. \model{} is supported by an innovative use of the formalism of random variables and complemented by rigorous mathematical proves.

The differentiable nature of \model{} makes it uniquely suited for integration with gradient-based optimization techniques. This capability enables the joint optimization of genetic quantities of interest and the performance of resulting agents in selected tasks, thus bridging a critical gap between biological plausibility and computing. Out validation experiments demonstrated the effectiveness of \model{} across multiple RL benchmarks, with agents derived from the framework consistently outperforming the considered baselines. These results highlight the potential of \model{} as a foundational tool for generating task-specific biological agents with unparalleled precision.

While \model{} marks significant advancements over the existing methodologies, important challenges remain before its application in real-world contexts can be conceptualized. Key gaps include the event-driven nature of biological neural networks, the employed synapse model, and the reliance on feedforward topologies that do not fully capture the complexity of biological wiring. We proposed several mitigations to address these limitations, such as adapting the framework for spiking neural networks, incorporating positional embeddings to account for distance-dependent connectivity, and leveraging experimental data to refine circuit-level properties. These extensions will not only enhance the biological plausibility of \model{} but also expand its applicability to more complex and realistic scenarios.

By enabling the precise manipulation of genetic rules governing neural circuit formation, we expect \model{} to lay the foundation for groundbreaking advancements in the development of biological agents and the realization of applications that are currently beyond our reach.


\section{Acknowledgments}

\begin{sloppypar}
This work is supported and funded by: \#NEXTGENERATIONEU (NGEU); the Ministry of University and Research (MUR); the National Recovery and Resilience Plan (NRRP); project MNESYS (PE0000006, to NT) - \textit{A Multiscale integrated approach to the study of the nervous system in health and disease} (DN. 1553 11.10.2022); the MUR-PNRR M4C2I1.3 PE6 project PE00000019 Heal Italia (to NT); the NATIONAL CENTRE FOR HPC, BIG DATA AND QUANTUM COMPUTING, within the spoke \textit{``Multiscale Modeling and Engineering Applications''} (to NT); the European Innovation Council (Project CROSSBRAIN - Grant Agreement 101070908, Project BRAINSTORM - Grant Agreement 101099355); the Horizon 2020 research and innovation Programme (Project EXPERIENCE - Grant Agreement 101017727). Tommaso Boccato is a PhD student enrolled in the National PhD in Artificial Intelligence, XXXVII cycle, course on Health and Life Sciences, organized by Università Campus Bio-Medico di Roma.
\end{sloppypar}

\bibliography{elife-sample}


\appendix
\begin{appendixbox}

\section{The Polarity Matrix}\label{sec:polarity}

As outlined in \citep{10.1371/journal.pcbi.1007974}, synapse polarity in \textit{C. elegans}, a well-studied small nematode, is described by the interplay among 3 neurotransmitters--glutamate, acetylcholine, and GABA--and their corresponding receptors. Specifically, each neurotransmitter can be associated with receptors capable of exerting both excitatory and inhibitory effects on synaptic connections. This relationship can be represented by a $3 \times (2 \cdot 3)$ polarity matrix:
\begin{equation}
    A = \begin{bmatrix}
        1 & -1 & 0 & 0 & 0 & 0\\
        0 & 0 & 1 & -1 & 0 & 0\\
        0 & 0 & 0 & 0 & 1 & -1
    \end{bmatrix}
\end{equation}
Here, each neurotransmitter synthesized in pre-synaptic neurons can be bind to either a positive (+) or negative (-) receptor in post-synaptic neurons. The 0s in the matrix signify that receptors attuned to a specific neurotransmitter are incapable of receiving different ones. This formalism readily extends to accommodate an arbitrary number of neurotransmitters by setting $M = 2L$ and expanding $A$ to an $L \times 2L$ block diagonal matrix, where each block is represented as $[1, -1]$. It is worth noting that, in the experiments described in Section \nameref{sec:results}, the entries of $A$ do not belong to the set of learnable parameters.

\section{Simulating Synaptogenesis}\label{sec:simulation}

Algorithm \ref{alg:simulation} outlines one possible implementation of the synaptogenesis simulation introduced at the end of Section \nameref{sec:methods}, which we used in our experiments.

\nolinenumbers
\begin{algorithm}[H]
\caption{Our synaptogenesis simulation procedure.}\label{alg:simulation}
    $\bar{G} \gets Q(A \odot K)R^T$\\
    $\tilde{B} \gets \bm{0}$\\
    \For{$i \gets 1$ \upshape{to} $G$}{
        \For{$j \gets 1$ \upshape{to} $G$}{
            $M \gets X_{:,i} \otimes X_{:,j}$\\
            $M \gets round(\alpha M)$\\
            $\tilde{B} \gets \tilde{B} + \begin{bmatrix}
                \dots & \dots & \dots\\
                \dots & \tilde{B}_{uv}^{ij} \sim \text{Bin}(M_{uv}, O_{ij}) & \dots\\
                \dots & \dots & \dots
            \end{bmatrix}$
        }
    }
    $W \gets \tilde{B} \odot \Big(\frac{1}{\alpha}\bar{G}\Big)$
\end{algorithm}

In this algorithm, $G$ represents the number of genes involved, notation $\bullet_{:,i}$ represents selection of the $i$-th column, $\otimes$ denotes the outer product, the method $round()$ performs pointwise rounding to the nearest integer, while $\tilde{B}_{uv}^{ij}$ is the contribution of gene pair $(i, j)$ to the number of synapses of neuron pair $(u, v)$. All quantities mentioned in Algorithm \ref{alg:simulation} are restricted to a specific neural layer.

The coefficient $\alpha$ is a general correction factor used to control the average degree of nodes in the generated networks. The necessity of this correction is discussed in Appendix \nameref{sec:quantization}. Here, we show that this correction does not impact the average weight matrices learned by \model{}:
\begin{proposition}\label{th:correction}
    The average weight matrix $\bar{W}^{corr.}$ resulting from the correction shown in Algorithm \ref{alg:simulation}, with $\alpha > 0$, coincides with the average weight matrix learned by \model{}, $\bar{W}$.
\end{proposition}
\begin{proof}
\begin{align*}
    \bar{W}^{corr.} &= \Big(\sqrt{\alpha}XOX^T\sqrt{\alpha}\Big) \odot \Big(\frac{1}{\alpha}\bar{G}\Big)\\
    &=(XOX^T) \odot \bar{G}\\
    &= \bar{W}
\end{align*}
\end{proof}

In all our simulations, we chose $\alpha$ such that the resulting networks had an average degree of $10^5$. While this value is quite high (the typical number of synapses per neuron in the human brain is on the order of magnitude of $10^4$ \citep{zhang2019basicneuralunitsbrain}), it aligns with the connectivity levels observed in certain types of cells in specific brain regions, e.g., Purkinje cells in the cerebellar cortex \citep{napper_number_1988}.

\section{A Problem of Quantization}\label{sec:quantization}

The synaptogenesis simulation introduces a subtle ``quantization'' issue. When the order of magnitude of the input provided to a \model{} network and the average conductances associated with each pair of neurons are fixed, the number of synapses determines the range of values each weight can take. Conversely, when the number of synapses is fixed, the average conductances dictate the granularity with which a given range can be represented. Thus, the interplay between these factors significantly affects the degree of synaptic weight quantization. Additionally, the role of rounding, introduced in Algorithm \ref{alg:simulation} to ensure that the parameters $n_{ij}$ of the binomial random variables are integers, cannot be overlooked. The inherently discrete nature of these random variables also plays a fundamental role in the quantization process.

It is therefore interesting to study the error $W - \bar{W}$, specifically the contribution to this error made by a single gene pair for a specific pair of neurons. It can be shown that:
\begin{proposition}\label{th:error}
    When considering an optimal simulated agent--characterized by a number of synapses as close as possible to the mean learned by \model{}--and assuming the correction introduced in Algorithm \ref{alg:simulation} is applied, the error $e$ between the mean number of synapses predicted by \model{} and the actual number of synapses in the agent is bounded by $\frac{|\bar{G}_{uv}|}{2\alpha}(p_{ij} + 1)$.
\end{proposition}
\begin{proof}
    \begin{align}
        e &= \Big|n_{ij}p_{ij}\cdot \bar{G}_{uv} - \text{round}(\text{round}(\alpha n_{ij})p_{ij}) \cdot \frac{1}{\alpha}\bar{G}_{uv}\Big|\label{eq:two-rounds}\\
        &\le \Big|\frac{1}{\alpha}\bar{G}_{uv}\Big| \cdot \Big|\alpha n_{ij}p_{ij} - \text{round}(\text{round}(\alpha n_{ij})p_{ij})\Big|\\
        &\le \Big|\frac{1}{\alpha}\bar{G}_{uv}\Big| \cdot \Big|\alpha n_{ij}p_{ij} - \text{round}\Big(\big(\alpha n_{ij} + \frac{1}{2}\big)p_{ij}\Big)\Big|\label{eq:round-le-1}\\
        &\le \Big|\frac{1}{\alpha}\bar{G}_{uv}\Big| \cdot \Big|\alpha n_{ij}p_{ij} - \Big(\big(\alpha n_{ij} + \frac{1}{2}\big)p_{ij} + \frac{1}{2}\Big)\Big|\label{eq:round-le-2}\\
        &= \frac{|\bar{G}_{uv}|}{2\alpha} (p_{ij} + 1)
    \end{align}
In this proof, $n_{ij}p_{ij} = \mathbb{E}[\text{Bin}(n_{ij}, p_{ij})]$, the outer $\text{round}$ in step \eqref{eq:two-rounds} helps account for the nearest integer approximation of the mean number of synapses learned by \model{}, and steps \eqref{eq:round-le-1} and \eqref{eq:round-le-2} are derived from bounds applied to the two rounding operations.\\
\end{proof}
It is evident that choosing a sufficiently large $\alpha$--and thus achieving a sufficiently high average neuronal degree--can bring the weight matrices of the best simulated agents closer to the optimal matrices identified by \model{}. This reduces the performance loss of these agents on the task for which they were pre-wired.

\section{Training Details}\label{sec:train-details}

This section provides all the details related to the training of the \model{} models that were not explicitly mentioned in Section \nameref{sec:methods}
Neural networks were implemented as custom MLPs featuring a single hidden layer with 128 neurons, a number that allowed us to easily conduct experiments with the computational resources available. The input and output layers were configured with a number of neurons equal to the dimensionality of the observation space and the number of possible actions in the selected RL environments, respectively. Consequently, the size of the input layer ranged from 2 to 8, while the size of the output layer ranged from 2 to 4. Details about the RL environments can be found in the Gym library documentation: \url{https://www.gymlibrary.dev/index.html}. Learnable parameters associated with genetic rules and conductances were initialized following the procedure described in \citep{barabasi_complex_2023}. All other parameters were initialized using a Kaiming normal distribution \citep{7410480}.

Trainings were conducted using the Adam optimizer \citep{DBLP:journals/corr/KingmaB14}, with its default parameters unchanged except for the learning rate. Training consisted of 500000 \textit{environment steps}. To minimize randomness in the results and mitigate sub-optimality associated with specific hyperparameter choices, the genetic profiles used in simulations were selected through a hyperparameter grid-search. This search explored three different learning rates (0.03, 0.003, 0.0003) and three random seeds for various sampling processes. Specifically, after hyperparameter optimization, we retained the checkpoint corresponding to the best validation performance achieved by the models. Validation was conducted every 10000 training steps. During both validation and subsequent simulations, agent rewards were computed as the average scores over 10 different episodes. The remaining hyperparameters related to the DQN algorithm were sourced from: \url{https://github.com/DLR-RM/rl-baselines3-zoo/blob/master/hyperparams/dqn.yml}. Additional information can be found in the released code.

\section{The Separable Natural Evolution Strategy (SNES) Baseline}\label{sec:snes}

As mentioned in Section \nameref{sec:results}, the \textit{SNES baseline} was constructed by replacing gradient descent in our \model{} framework with SNES, an optimization technique introduced in \citep{10.1145/2330784.2330815} and employed by the authors of \citep{Stockl2021.05.18.444689} for a related task, namely the optimization of \textit{probabilistic skeletons}. The core idea behind SNES is to sample $\lambda$ \textit{offsprings} from a normal distribution of the type $\mathcal{N}(\bm{\mu},I\bm{\sigma})$, where $\bm{\mu}$ is built by concatenating the flattened learnable matrices of \model{}. These offsprings are evaluated based on their \textit{fitness}, allowing the distribution to adapt and better capture those regions of the parameter space where offspring fitness is higher. The implementation details of SNES are provided in the pseudo-code of Algorithm \ref{alg:snes}.

\nolinenumbers
\begin{algorithm}[H]
\caption{The SNES optimization technique.}\label{alg:snes}
num\_steps $\gets 0$\\
max\_steps $\gets 500000$\\
\While{\upshape{num\_steps} $<$ \upshape{max\_steps}}{
    \For{$k \gets 1$ \upshape{to} $\frac{\lambda}{2}$}{
        $\bm{s}_k \sim \mathcal{N}(\bm{0}, I)$\\
        $\bm{s}_{k + \frac{\lambda}{2}} \gets -\bm{s}_k$
    }
    \For{$k \gets 1$ \upshape{to} $\lambda$}{
        $\bm{\theta}_k \gets \bm{\mu} + \bm{\sigma} \odot \bm{s}_k$
    }
    $\nabla_{\bm{\mu}} \gets \sum_{k = 1}^\lambda F(\bm{\theta}_k)\bm{s}_k$\\
    $\nabla_{\bm{\sigma}} \gets \sum_{k = 1}^\lambda F(\bm{\theta}_k)(\bm{s}_k^T\bm{s}_k - 1)$\\
    $\bm{\mu} \gets \bm{\mu} + \eta_{\bm{\mu}}\bm{\sigma}\nabla_{\bm{\mu}}$\\
    $\bm{\sigma} \gets \bm{\sigma}e^{\frac{\eta_{\bm{\sigma}}}{2}\nabla_{\bm{\sigma}}}$\\
    num\_steps += $steps\_performed()$
}
\end{algorithm}

Inside the algorithm, the vectors $\bm{s}_k$ contain the ``noise'' used in generating the offsprings, $\bm{\sigma}$ is a vector of variances initialized in accordance with \citep{Stockl2021.05.18.444689}, $F$ is the fitness function, $\eta_{\bm{\mu}}$ and $\eta_{\bm{\sigma}}$ are two learning rates $> 0$, and the method $steps\_performed()$ returns the number of actions taken during the current iteration on the RL environment being optimized. Specifically, we defined the fitness function $F(\bullet)$ as the average performance achieved by $m$ simulated biological agents based on the genetic profile provided as input to the function. Individual scores were calculated by averaging the rewards obtained by an agent over 10 different episodes.

To ensure fair comparisons with the profiles obtained through gradient descent, we limited the number of actions executable on a RL environment to 500000 in the context of a single optimization. Additionally, for this baseline, we conducted a hyperparameter search similar to the one described in Appendix \nameref{sec:train-details}, varying $m$ in the set $\{10, 20, 30\}$ and assigning $\lambda$, during the search, the values 8, 16, and the default value of the SNES implementation used (\url{https://github.com/pybrain/pybrain}).

\section{The Bio-Plausible Baseline}\label{sec:bio-plausible-init}

As a control, we designed an additional baseline, referred to as \textit{bio-plausible} in the main text, intended to model a scenario where a neural population with a predefined number of neurons is left free to form synapses without any manipulation of genes or gene expression in individual neurons tailored to a specific task. In other words, this baseline was designed to assess whether there is a performance difference between agents produced by \model{} and neuronal networks that develop without  guidance, or equivalently, to determine whether the scores achieved by \model{} agents are due to chance.

To do this, we fixed the genetic rules learned by \model{} for each environment as the biological ground truth and initialized the gene expression of the individual neurons in the tested agents according to the \textit{lineal model} proposed in \citep{kerstjens_constructive_2022}. This model is simple yet profound and elegant, and it is capable of predicting phenomena later validated by existing experimental data \citep{Kerstjens2024.09.19.613507}. According to the lineal model, the expression inherited by two cells $u$ and $v$ resulting from mitosis is calculated by adding a differential expression vector, drawn from a normal distribution, to the expression of the parent cell $p$. Formally:
\begin{align}
    &\bm{c}_u = \bm{c}_p + \bm{\delta}_u\\
    &\bm{c}_v = \bm{c}_p + \bm{\delta}_v\\
    &\bm{\delta}_u, \bm{\delta}_v \sim \mathcal{N}\big(\bm{0}, I\big)
\end{align}
Here, the genetic expression pattern $\bm{c} = [\bm{x}^T, \bm{q}^T, \bm{r}^T]^T$ of a generic neuron is understood as the concatenation of genetic expression responsible for synaptic multiplicity, neurotransmitter distribution, and receptor distribution. With a slight abuse of notation, we are actually using these variables to refer to the previously mentioned vectors before they are normalized or otherwise mapped into the domain of the genetic quantities of interest.

In more detail, for each agent, we sampled the genetic expression profile of a virtual zygote from a normal distribution and simulated a lineage tree with as many leaves as the number of neurons in the agent. This allowed us to assign gene expression profiles to individual cells following the equations above. We finalized the procedure by randomly assigning each neural layer expression profiles corresponding to leaves from the same complete subtree, simulating spatial contiguity for neurons within the same layer. The procedure we implemented is detailed in Algorithm \ref{alg:lineal-model}. In the pseudo-code, the first line initializes the zygote’s genetic expression, $N$ corresponds to the target number of cells to be produced, the notation $\bullet_{1\dots N}$ selects the first $N$ columns of the respective matrix, function $rand\_roll()$ circularly permutes the columns of the input matrix, and $C^{in}$, $C^{hidden}$ and $C^{out}$ represent the matrices associated with the genetic expressions of the various layers of the agent being initialized.

\nolinenumbers
\begin{algorithm}[H]
\caption{Our \textit{lineal model}-based initialization procedure for gene expression.}\label{alg:lineal-model}
$C \gets [\bm{c} \sim \mathcal{N}(\bm{0}, I)]$\\
num\_cells $\gets 1$\\
\While{\upshape{num\_cells} $< N$}{
    $C^{new} \gets [\ ]$\\
    \For{$\bm{c} \in C$}{
       $\bm{\delta}_u, \bm{\delta}_v \sim \mathcal{N}\big(\bm{0}, I\big)$\\
       $\bm{c}_u = \bm{c} + \bm{\delta}_u$\\
       $\bm{c}_v = \bm{c} + \bm{\delta}_v$\\
       $C^{new} \gets [C^{new}, \bm{c}_u, \bm{c}_v]$
    }
    $C \gets C^{new}$\\
    num\_cells $\gets$ $2$num\_cells
}
$C \gets C_{1\dots N}$\\
$C \gets rand\_roll(C)$\\
$[C^{in}, C^{hidden}, C^{out}] \gets C$
\end{algorithm}

\section{Reward Thresholds for Considering a Task Solved}\label{sec:resolution}

There is no consensus within the RL community on the definition of ``solving'' when it comes to the most popular benchmark tasks used for evaluating new algorithms. Therefore, we derived reward thresholds, based on the guidelines provided at \url{https://www.gymlibrary.dev}, to serve as a reference for distinguishing agents capable of solving their respective RL environments from those unable to achieve satisfactory performance.

For Cart Pole, the threshold was set to 195, which represents the number of consecutive time steps an agent must keep the pole balanced. The environment provides a reward of +1 for every step the pole remains upright. For Mountain Car, we used a threshold of -200. The objective of this task is to propel the car to the top of a hill within the environment. We consider the task solved in all cases where the car successfully reaches the goal, and consider it unsolved only when the car fails to reach the goal within the simulation time, which results in a reward of -200. In Lunar Lander, the threshold is set to 200, a reward that Gym documentation associates with a soft landing on the designated landing pad, without any crashes. Finally, for Acrobot, we consider the task solved if the agent manages to swing the chain above the target height before the simulation ends. This corresponds to achieving a reward greater than -500.

\section{Additional Results}\label{sec:additional-results}

We provide here, for the sake of completeness, the results obtained by optimizing models characterized by genetic profiles containing 32 and 64 genes. These results are shown in Figures \ref{fig:reward-dist-32}, \ref{fig:reward-dist-64} and Tables \ref{tab:aggregated-32}, \ref{tab:aggregated-64}.

\begin{center}
    \includegraphics[width=\textwidth]{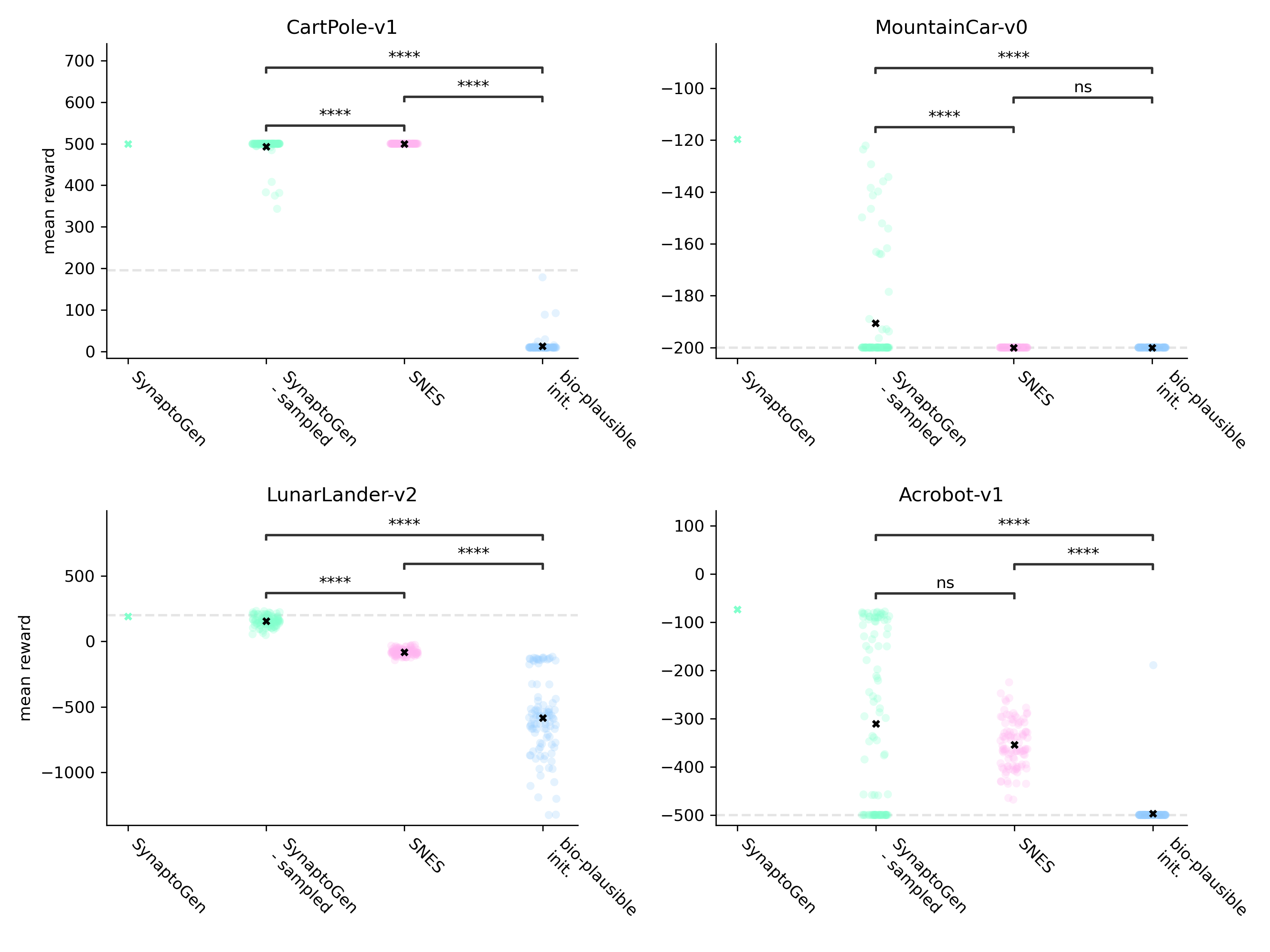}
    \captionof{figure}{\resfigcaption{32}}
    \label{fig:reward-dist-32}
\end{center}

\begin{center}
    \small
    \begin{tabular}{llrrrrr}
    \toprule
     &  & \textbf{mean} & \textbf{std} & \textbf{top-10 mean} & \textbf{top-10 std} & \textbf{\% solved} \\
    \textbf{environment} & \textbf{model} &  &  &  &  &  \\
    \midrule
    \multirow[c]{3}{*}{CartPole-v1} & SynaptoGen & 493.31 & 27.04 & \textbf{500.00} & 0.00 & \textbf{100.00\%} \\
     & SNES & \textbf{500.00} & 0.00 & \textbf{500.00} & 0.00 & \textbf{100.00\%} \\
     & bio-plausible & 13.19 & 20.32 & 47.28 & 55.74 & 0.00\% \\
    \cmidrule{2-7}
    \multirow[c]{3}{*}{MountainCar-v0} & SynaptoGen & \textbf{-190.63} & 20.93 & \textbf{-136.09} & 9.09 & \textbf{22.00\%} \\
     & SNES & -200.00 & 0.00 & -200.00 & 0.00 & 0.00\% \\
     & bio-plausible & -200.00 & 0.00 & -200.00 & 0.00 & 0.00\% \\
    \cmidrule{2-7}
    \multirow[c]{3}{*}{LunarLander-v2} & SynaptoGen & \textbf{154.50} & 39.34 & \textbf{217.98} & 8.39 & \textbf{17.00\%} \\
     & SNES & -81.45 & 24.69 & -36.21 & 6.18 & 0.00\% \\
     & bio-plausible & -583.41 & 293.58 & -128.53 & 5.40 & 0.00\% \\
    \cmidrule{2-7}
    \multirow[c]{3}{*}{Acrobot-v1} & SynaptoGen & \textbf{-311.14} & 180.04 & \textbf{-80.78} & 1.56 & 61.00\% \\
     & SNES & -354.02 & 49.73 & -268.30 & 21.28 & \textbf{100.00\%} \\
     & bio-plausible & -496.89 & 31.06 & -468.94 & 98.22 & 1.00\% \\
    \bottomrule
    \end{tabular}

    \captionof{table}{\restabcaption{32}}
    \label{tab:aggregated-32}
\end{center}

\begin{center}
    \centering
    \includegraphics[width=\textwidth]{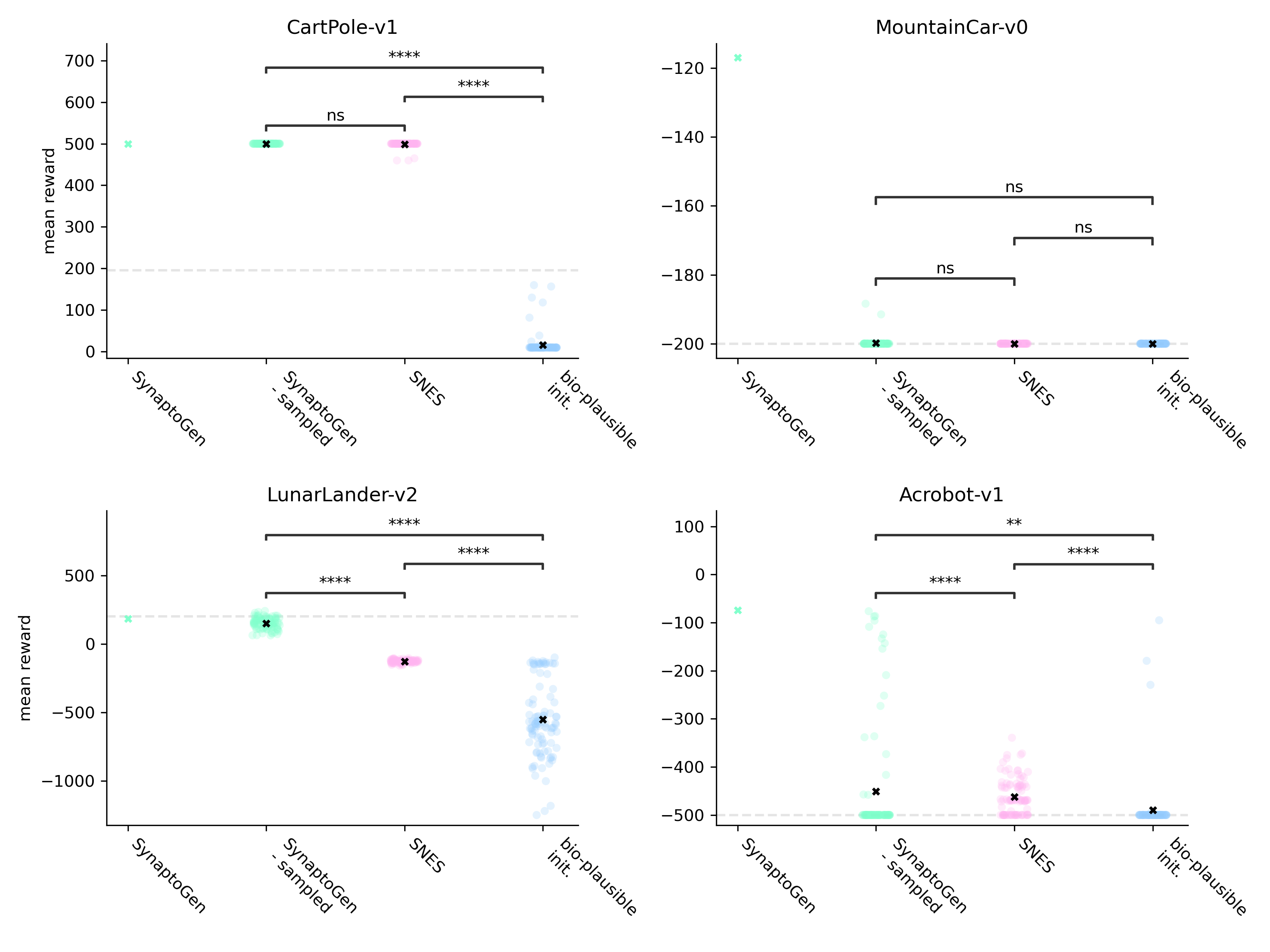}
    \captionof{figure}{\resfigcaption{64}}
    \label{fig:reward-dist-64}
\end{center}

\begin{center}
    \small
    \begin{tabular}{llrrrrr}
    \toprule
     &  & \textbf{mean} & \textbf{std} & \textbf{top-10 mean} & \textbf{top-10 std} & \textbf{\% solved} \\
    \textbf{environment} & \textbf{model} &  &  &  &  &  \\
    \midrule
    \multirow[c]{3}{*}{CartPole-v1} & SynaptoGen & \textbf{500.00} & 0.00 & \textbf{500.00} & 0.00 & \textbf{100.00\%} \\
     & SNES & 498.84 & 6.63 & \textbf{500.00} & 0.00 & \textbf{100.00\%} \\
     & bio-plausible & 15.93 & 27.03 & 75.00 & 61.07 & 0.00\% \\
    \cmidrule{2-7}
    \multirow[c]{3}{*}{MountainCar-v0} & SynaptoGen & \textbf{-199.80} & 1.43 & \textbf{-197.99} & 4.30 & \textbf{2.00\%} \\
     & SNES & -200.00 & 0.00 & -200.00 & 0.00 & 0.00\% \\
     & bio-plausible & -200.00 & 0.00 & -200.00 & 0.00 & 0.00\% \\
    \cmidrule{2-7}
    \multirow[c]{3}{*}{LunarLander-v2} & SynaptoGen & \textbf{151.21} & 40.20 & \textbf{215.45} & 13.43 & \textbf{13.00\%} \\
     & SNES & -127.01 & 10.55 & -109.01 & 3.63 & 0.00\% \\
     & bio-plausible & -552.04 & 270.15 & -129.60 & 12.52 & 0.00\% \\
    \cmidrule{2-7}
    \multirow[c]{3}{*}{Acrobot-v1} & SynaptoGen & \textbf{-451.23} & 118.92 & \textbf{-121.76} & 40.33 & 18.00\% \\
     & SNES & -462.01 & 37.41 & -385.67 & 21.67 & \textbf{68.00\%} \\
     & bio-plausible & -489.99 & 57.73 & -399.88 & 163.21 & 4.00\% \\
    \bottomrule
    \end{tabular}

    \captionof{table}{\restabcaption{64}}
    \label{tab:aggregated-64}
\end{center}

\section{nDGE-based Bio-Plausible Rules}\label{sec:bio-rules}

To produce the genetic rules used in the bio-plausible validation in Section \nameref{sec:results}, we first computed the gene co-expressions of the \textit{C. elegans} nerve ring by performing a generalized nDGE analysis on the dataset provided in \citep{TAYLOR20214329}. This generalization was necessary because the nDGE implementation available at \url{https://github.com/cengenproject/connectivity_analysis} is designed to calculate local co-expressions, specifically limited to a set of cells consisting of a pre-synaptic neuron and its corresponding post-synaptic partners. 

\begin{sloppypar}
To compute global co-expressions, we identified all pairs of neurons that shared synapses and all the pairs that were in contact without being connected. Then, for each possible pair of genes, we calculated co-expressions for both sets and conducted a T-test with Bonferroni correction. Pairs with a $p$-value $< 0.05$ were considered co-expressed. The detailed procedure is outlined in Algorithm \ref{alg:global-ndge}. In this pseudo-code, $B$ and $C$ represent the connectome and the contactome of the neuronal network under analysis, respectively, $\oplus$ denotes a pointwise XOR operation, and $G$ corresponds to the number of genes analyzed. The third-level ``for'' loops, instead, separate the indices of pre- and post-synaptic neurons involved in the pairs being evaluated in each iteration. The co-expression of a specific pair of genes for all neuron pairs in a given set is calculated as the pointwise product $C_{i,\bm{u}} \odot C_{j,\bm{v}}$, where $C$ is the matrix containing gene expression data for all neurons, and the notation $\bullet_{i,\bm{u}}$ selects the $i$-th row (corresponding to gene $i$) and the columns associated with the neurons in $\bm{u}$. The method $ttest()$ performs the T-test, while $p_{ij}$ and $g_{ij}$ represent the $p$-value for the gene pair $(i, j)$ and its corresponding binary co-expression, respectively.
\end{sloppypar}

\nolinenumbers
\begin{algorithm}[H]
\caption{Our global nDGE variant.}\label{alg:global-ndge}
with\_synapses $\gets \{(u, v): B_{uv} = 1\}$\\
contact\_only $\gets \{(u, v): C_{uv} \oplus B_{uv} = 1\}$\\
\For{$i \gets 1$ \upshape{to} $G$}{
    \For{$j \gets 1$ \upshape{to} $G$}{
        $\bm{u}, \bm{v} \gets [], []$\\
        \For{$(u, v) \in$ \upshape{with\_synapses}}{
            $\bm{u} \gets [\bm{u}^T, u]^T$\\
            $\bm{v} \gets [\bm{v}^T, v]^T$
        }
        coexp\_ws $\gets \ln(\bm{1} + C_{i,\bm{u}}\odot C_{j,\bm{v}})$\\
        $\bm{u}, \bm{v} \gets [], []$\\
        \For{$(u, v) \in$ \upshape{contact\_only}}{
            $\bm{u} \gets [\bm{u}^T, u]^T$\\
            $\bm{v} \gets [\bm{v}^T, v]^T$
        }
        coexp\_co $\gets \ln(\bm{1} + C_{i,\bm{u}}\odot C_{j,\bm{v}})$\\
        $p_{ij} \gets ttest($coexp\_ws, coexp\_co$)$\\
        \uIf{$G^2p_{ij} < 0.05$}{
            $g_{ij} \gets 1$
        }
        \Else{
            $g_{ij} \gets 0$
        }
    }
}
\end{algorithm}

\begin{sloppypar}
After obtaining the co-expressions (illustrated in Figure \ref{fig:bio-rules}), and to avoid constructing overly artificial rules--since co-expression reflects correlation while genetic rules imply causation--while still maintaining sufficient bio-plausibility, we allowed the models to learn the matrix $O$ but constrained the domain of its entries. Specifically, starting from \model{}'s learnable parameters, we computed the genetic rules matrix $O$ differently for co-expressed and non-co-expressed gene pairs. The approach involved pushing the rules for co-expressed pairs toward probabilities close to 1, while those for non-co-expressed pairs were pushed toward probabilities close to 0. This was implemented using two sigmoid functions, also shown in Figure \ref{fig:bio-rules}:
\begin{equation}\label{eq:sigs}
    O_{ij} = \begin{cases}
        \frac{1}{2}\text{sigmoid}\bigg(\frac{\hat{O}_{ij} + 3\sigma_{\hat{O}}}{T}\bigg) + \frac{1}{2},\quad &\text{if}\ g_{ij} = 1\\[1em]
        \frac{1}{2}\text{sigmoid}\bigg(\frac{\hat{O}_{ij} - 3\sigma_{\hat{O}}}{T}\bigg),\quad &\text{if}\ g_{ij} = 0       
    \end{cases}   
\end{equation}
Here, $\hat{O}$ represents the parameter matrix from which the genetic rules $O$ are derived, $\sigma_{\hat{O}}$ is the standard deviation of the $\hat{O}$ values' distribution, and $T$ is the temperature used. The learned genetic rule matrices shown in Figure \ref{fig:bio-rules} confirm that \model{} assigned high probabilities (i.e., approximately 1) to co-expressed pairs, supporting the utility of co-expression data for optimization purposes.
\end{sloppypar}

\end{appendixbox}

\end{document}